\documentclass[final]{cvpr}

\usepackage{times}
\usepackage{epsfig}
\usepackage{graphicx}
\usepackage{amsmath}
\usepackage{amssymb}

\usepackage{algorithm}  
\usepackage[noend]{algorithmic} 
\usepackage{multirow}
\usepackage{subcaption}
\usepackage{pifont}

\usepackage[pagebackref=true,breaklinks=true,colorlinks,bookmarks=false]{hyperref}

\def\cvprPaperID{3454} 
\def\confYear{CVPR 2021}

\newcommand{\cmark}{\ding{51}}%
\newcommand{\xmark}{\ding{55}}%

\begin{document}

\title{Optimal Gradient Checkpoint Search for Arbitrary Computation Graphs}

\author{Jianwei Feng and Dong Huang\\ 
Robotics Institute, 
Carnegie Mellon University\\
Pittsburgh, PA 15213 \\
jfeng1@andrew.cmu.edu, donghuang@cmu.edu 
}


\maketitle

\newtheorem{Def}{Definition} 
\newtheorem{Theo}{Theorem}
\newtheorem{Lemma}{Lemma}

\begin{abstract}
Deep Neural Networks(DNNs) require huge GPU memory when training on modern image/video databases. Unfortunately, the GPU memory is physically finite, which limits the image resolutions and batch sizes that could be used in training for better DNN performance. Unlike solutions that require physically upgrade GPUs, the Gradient CheckPointing(GCP) training trades computation for more memory beyond existing GPU hardware. GCP only stores a subset of intermediate tensors, called Gradient Checkpoints (GCs), during forward. Then during backward, extra local forwards are conducted to compute the missing tensors. The total training memory cost becomes the sum of (1) the memory cost of the gradient checkpoints and (2) the maximum memory cost of local forwards. To achieve maximal memory cut-offs, one needs optimal algorithms to select GCs. Existing GCP approaches rely on either manual input of GCs or heuristics-based GC search on Linear Computation Graphs (LCGs), and cannot apply to Arbitrary Computation Graphs(ACGs). In this paper, we present theories and optimal algorithms on GC selection that, for the first time, are applicable to ACGs and achieve the maximal memory cut-offs. Extensive experiments show that our approach not only outperforms existing approaches (only applicable on LCGs), and is applicable to a vast family of LCG and ACG networks, such as Alexnet, VGG, ResNet, Densenet, Inception Net and highly complicated DNNs by Network Architecture Search. Our work enables GCP training on ACGs, and cuts off up-to 80\% of training memory\footnote{Cutting off 80\% of training memory means one can double the input image size or quadruple the batch size on the same GPUs.} with a moderate time overhead ($\sim$ 30\%-50\%).  Codes are available\footnote{ \href{https://github.com/lordfjw/OptimalGradCheckpointing}{https://github.com/lordfjw/OptimalGradCheckpointing}}.

\end{abstract}

\section{Introduction}
\begin{figure}[t]
\centering
\includegraphics[width=.95\linewidth]{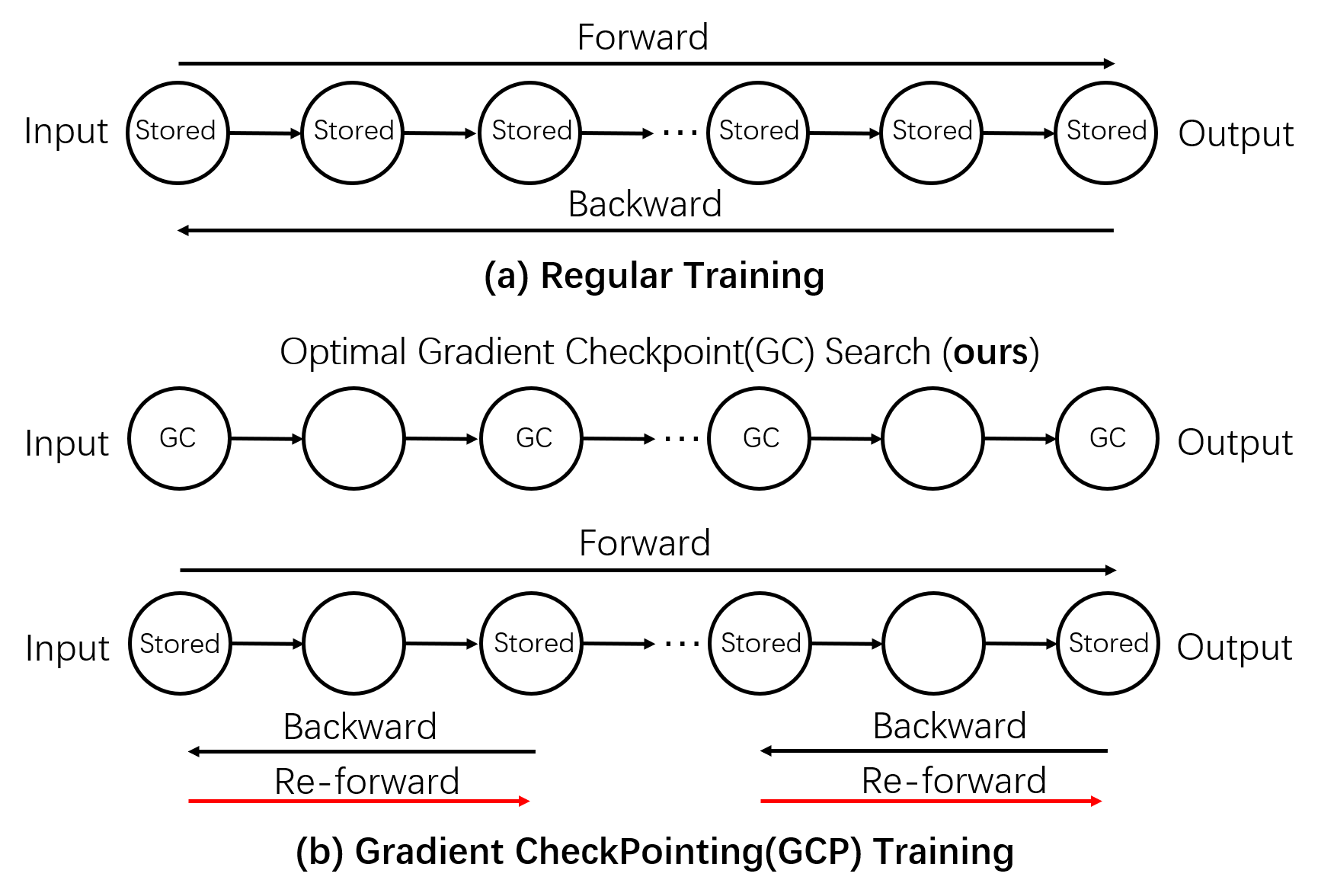}\hspace{-.3cm}
\includegraphics[width=.9\linewidth]{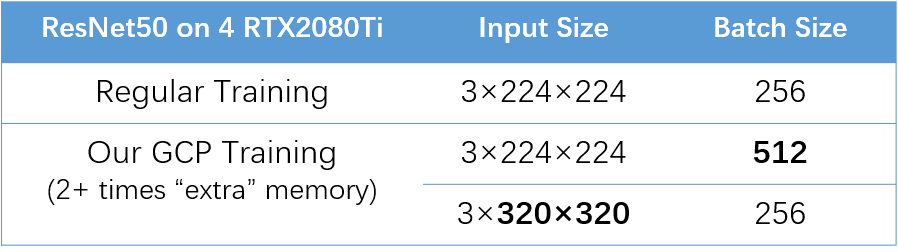}\hspace{-.3cm}
\caption{Regular Training vs. Gradient CheckPointing(GCP) Training. (a) The regular training stores all tensors during forward, and uses these tensors to compute gradients during backward. (b) GCP stores a subset of tensors during the first forward, and conducts extra local re-forwards to compute tensors and gradients during backward. Our approach automatically searches the optimal set of Gradient Checkpoints (GCs) for memory cut-off. Such that on the same \textbf{physical} GPU memory (e.g., in 4 RTX2080Ti GPUs), GCP training can accommodate models that require $2+$ times extra GPU memory. }
\label{FigCompare}
\end{figure}
Deep Neural Networks(DNNs) require huge GPU memory when training on modern image/video databases. For popular backbone DNNs used for image feature extraction, such as AlexNet \cite{krizhevsky2012imagenet}, VGG \cite{simonyan2014very} and ResNet \cite{he2016deep}, the memory cost increases quadratically with the input image resolution and network depth. For example, given a median size input tensor of $\tiny{[BatchSize \times Channel \times Width\times Height]}=[32,3,224,224]$, ResNet101 requires  around $4$ GB memory only to store feature tensors and gradients in training (software overheads not included). In more challenging tasks, DNNs that detect small objects and large number of object categories require input image resolution of more than $600\times 600$ \cite{ren2015faster,Singh2017,RedmonFarhadi2018} and can easily consume more than $30$ GB with the same batch size. The memory issue is even worse for video-based DNNs, such as CDC \cite{shou2017cdc}, C3D \cite{ji20133d} and 3D-ResNet \cite{hara2017learning}. To recognize complex activities in video, the input video clips would be as long as $64$ frames and could easily go beyond $10$ GB using a moderate network. Memory issue also occurs in training DNN compositions, such as Generative Adversarial Networks (GANs), where multiple generator and discriminator networks are simultaneously stored in GPU memory. 

Existing efforts to address memory issues presented three main approaches: (1) Better single GPUs. Recent GPUs provide larger memory at the expense of exponentially growing price and power consumption. For instance, from TitanXp, Quadro P6000, RTX 3090 to Tesla V100, for 1-2.7 times increase in memory, the prices increase 2.8-8.5 times.  (2) Parallelization among multiple GPUs \cite{dean2012large,shi2009hash,langford2009slow,mcdonald2009efficient,mcdonald2010distributed,zinkevich2010parallelized,agarwal2014reliable,agarwal2011distributed}, which requires expensive clusters, introduces substantial I/O cost, and does not reduce the total memory cost. (3) Gradient CheckPointing (GCP) \cite{chen2016training,gruslys2016memory}, which focuses on trading computation for memory and reduces the total memory cost without any upgrade in hardware. Note that recent affordable GPUs (e.g., RTX 2080 Ti , RTX 3080), although limited in memory (around 11GB), provide exceptional improvement in GPU cores and FLOPS. Trading computation costs for memory is a very attractive solution that make it possible to train very heavy DNNs with finite GPU memory.

The regular DNN training approach consists of two alternated stages: forward and backward. Fig.~\ref{FigCompare} (a) illustrates an example of feed-forward neural networks. In the forward stage, the network takes an input tensor, and computes tensors at every layer to the final output. In the backward stage, the difference between the output and ground truth is passed back along the network to compute the gradients at each layer. The regular training approach saves tensors at all layers computed during forward. The total memory cost is the sum of cost over all these intermediate tensors.

GCP is a high-level training approach that trade extra computation time for substantial saving of GPU memory. Fig.~\ref{FigCompare} (b) illustrates its main idea. During GCP training, only a subset of intermediate tensors (which are called Gradient Checkpoints (GCs)) are stored in the first forward, and the missing tensors needed during backward are computed via extra local re-forwards. The total memory cost is the sum of the cost at the subset of intermediate tensors and the maximum memory cost among local re-forwards. Training with GCP can lead to substantial memory reduction, with the time overhead of local re-forwards. To achieve maximal memory cut-offs, one needs optimal algorithms to search for GCs. \textbf{The GC searching algorithm is a preprocessing step of GCP training, and only needs to be run once for one computation graph.} 

In this paper, we propose sophisticate theories and efficient algorithms that, for the first time, automatically find the \textbf{optimal} GCs in \textbf{Arbitrary} Computation Graphs(ACG), opens the gate of GCP training to a vast family of DNNs from ResNet to the Neural Architecture Search(NAS) networks.  Compared to existing GC searching (only applicable to Linear Computation Graphs(LCG) such as VGG), the optimality of our approach does not pose any assumption on computation graph, thus applicable to ACGs. Our optimal GCs lead to the smallest memory cost in GCP training. Using our GC searching algorithm, the GCP training can accommodate much larger models, on the same \textbf{physical} GPU memory (see the table in Fig.~\ref{FigCompare}). For instance, on 4 RTX2080Ti GPUs, regular training can typically train a ResNet50 image classification model of $3\times224\times224$ input size with $256$ batch size.

\section{Related Work}
To alleviate the memory pressure from a single GPU processor, many researchers utilized the well-established techniques for distributed computation \cite{dean2012large,shi2009hash,langford2009slow,mcdonald2009efficient,mcdonald2010distributed,zinkevich2010parallelized,agarwal2014reliable,agarwal2011distributed}. These techniques distribute memory pressure to possibly infinite GPUs or server clusters, but do not reduce the total memory cost of DNNs.

Some researchers reduced the memory usage by optimizing computation graph of DNN and performing liveness analysis. The computation graph of DNNs describes the dependencies of tensors among layers. Liveness analysis recycles garbage to manage memory. These ideas were originated from compiler optimization \cite{aho1986compilers} and has been widely adopted by deep learning frameworks: Theano \cite{bastien2012theano,bergstra2010theano}, MXNet \cite{chen2015mxnet}, Tensorflow \cite{abadi2016tensorflow} and CNTK \cite{yu2014introduction}. Some other techniques efficiently swap data between CPU and GPU \cite{wang2018superneurons,rhu2016vdnn}. These techniques usually cost extra I/O time and still do not actually reduce the total memory cost.

Other approaches focus on trading computation for memory with the idea of Gradient CheckPointing. Popular deep learning frameworks such as Pytorch \cite{paszke2017automatic} and Tensorflow \cite{abadi2016tensorflow} provide functions for users to manually define GCs in computation graph and perform gradient checkpoint training. These functions are user-dependent and their performance highly relies on the selected GCs. 

There are also algorithms to search for GCs automatically. Early in 2000, Griewank and Walther\cite{griewank2000algorithm} propose an optimal algorithm for linear computation graph assuming identical memory cost for each layer. Later, Chen et al. \cite{chen2016training} develop a greedy algorithm to search for GCs for linear computation graph (LCG), based on a heuristic that each segment has similar memory cost. Chen's algorithm is also adopted by OpenAI \cite{openai2018}. However, it is only applicable and not optimal for linear computation graph, and it's not applicable for non-linear computation graph, such as Inception net \cite{szegedy2016rethinking} and Dense net \cite{huang2017densely}.

Gruslys et al. \cite{gruslys2016memory} targets at Gradient CheckPointing for recurrent neural network (RNN). In recurrent neural network, the hidden state of each time step has the same size and thus has identical memory cost. Gruslys utilizes this characteristic and develops dynamic programming algorithm to solve for optimal GCs for RNN given a memory budget. Gruslys's approach is restrictive to RNN and can not generalize to network with arbitrary computation graph.

The main contribution of this paper is proposing algorithms to solve \textbf{optimal} GCs for \textbf{arbitrary computation graph (ACG)}. The difference between our approach and other approaches is summarized in Table.\ref{TabCompare}.
\begin{table*}
\caption{\cmark\cmark is both applicable and optimal, \cmark\xmark is applicable but not optimal, \xmark\xmark is not applicable nor optimal.}
\label{TabCompare}
\centering

\setlength{\tabcolsep}{1pt}
\fontsize{8}{10}\selectfont
\begin{tabular}{|c|c|c|c|c|c|}
\hline
Approach&\begin{tabular}{c} applicable \& optimal in \\identical cost LCG\end{tabular} & \begin{tabular}{c} applicable \& optimal in \\arbitrary cost LCG \end{tabular}&  \begin{tabular}{c} applicable \& optimal in \\ACG \end{tabular} &\begin{tabular}{c} automatic \end{tabular} &\begin{tabular}{c} with budget\end{tabular}\\
\hline
manual input&\cmark\xmark&\cmark\xmark&\cmark\xmark&\xmark&\xmark\\
Griewank\&Walther's\cite{griewank2000algorithm}&\cmark\cmark&\xmark\xmark&\xmark\xmark&\cmark&\xmark\\
Chen's\cite{chen2016training}&\cmark\cmark&\cmark\xmark&\xmark\xmark&\cmark&\xmark\\
Gruslys's\cite{gruslys2016memory}&\cmark\cmark&\xmark\xmark&\xmark\xmark&\cmark&\cmark\\
ours&\cmark\cmark&\cmark\cmark&\cmark\cmark&\cmark&\xmark\\

\hline
\end{tabular}
\end{table*}

\section{Overview}

GCP training consists of a pre-processing and a training step. In the pre-processing step, a GC searching algorithm is run to select GCs. Then in the training step, only tensors at the GCs are stored in memory during the first forward. During backward, the missing tensors and gradients are recovered by local re-forwarding. Like other GC searching algorithm \cite{griewank2000algorithm,chen2016training,gruslys2016memory}, \textbf{our algorithms focus on solving optimal GCs in the pre-processing step and is thus an one-time effort, which is only conducted before training}.

In section 4, we start with the Linear Computation Graph (LCG) and formulate the optimization problem of solving GCs. We first discuss a special case of LCGs, where we can easily compute an optimal solution in analytic form and understand the effectiveness of GCP. Then we present our algorithms to solve for optimal GCs in arbitrary LCGs.

In section 5, we present our approach on Arbitrary Computation Graphs (ACGs). We first introduce all the basic components, including definitions and sub-algorithms, and then the final solver based on these components.

In section 6, we present extensive experiments on networks with both linear and non-linear computation graphs. Due to space limit, we cannot put all illustrative examples in the paper. Extra illustrative examples are included in the "Extra Examples" section of the supplementary material.

In section 7, we present our conclusion for this paper.

\section{Linear Computation Graph (LCG)}
We denote a computation graph of a DNN as an acylic directed graph $G=\big(E, V\big)$. $E=\{e_i\}$ and $V=\{v_i\}$ are the edges and vertices in the graph respectively. The vertices represent the intermediate tensors and the edges represent DNN operations, such as convolution, matrix multiplication, etc. We denote function $l(\cdot)$ as a measure of memory cost. In practice for a single tensor $v_i$, $l(v_i)$ can be measured by the size of the tensor. We denote $V^R$ as the subset of vertices selected as GCs. $l(v_i^R)$ is defined as the memory cost of the $i$th gradient checkpoint in $V^R$. For two adjacent gradient checkpoint $v_i^R$ and $v_{i+1}^R$ in set $V^R$, suppose the ith gradient checkpoint $v_i^R$ corresponds to vertex $v_j$ in the original computation graph, and $v_{i+1}^R$ corresponds to $v_k$, the memory cost during re-forwards from $v_i^R$ to $v_{i+1}^R$ is defined as $l(v_i^R, v_{i+1}^R)=\sum_{t=j+1}^{k-1} l(v_t)$, which is the sum of cost over all the vertices between $v_j$ and $v_k$ in the computation graph. Using these notations, solving the optimal GCs is formulated as an optimization problem:
\begin{equation}
\min\limits_{V^R}( \sum_i l(v_i^R) + \max \limits_{i} l(v_i^R, v_{i+1}^R)), \label{eqnCostReforward}
\end{equation}
where the $\sum_i l(v_i^R)$ is the sum of the memory cost over all the GCs, and $\max \limits_{i} l(v_i^R, v_{i+1}^R))$ is the maximal cost among the local re-forwards. Eqn.~\ref{eqnCostReforward} describes the peak memory during gradient checkpoint training. Solution to Eq.~\ref{eqnCostReforward} produces the optimal GCs in $V^R$. 

\begin{figure}[h]
\centering
\includegraphics[width=.99\linewidth]{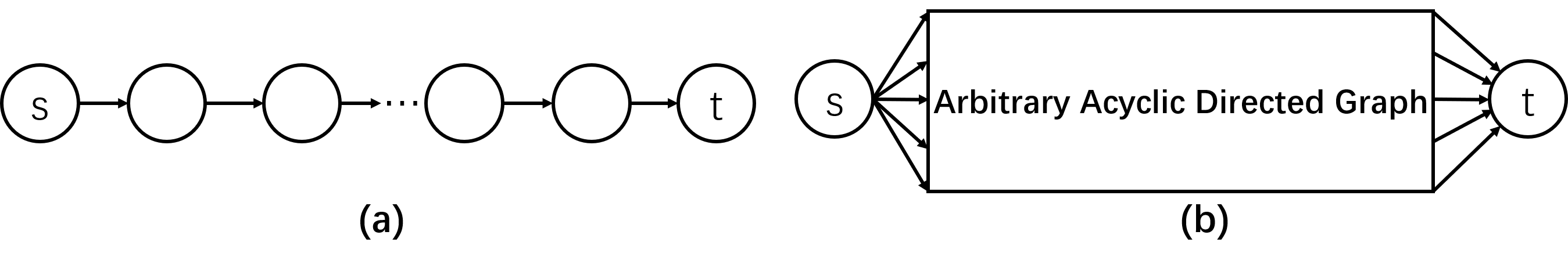}\hspace{-.1cm}
\caption{(a) Linear Computation Graph (LCG). ``s'' denotes the source vertex,``t'' denotes the target vertex.  (b) Arbitrary Computation Graph (ACG). The structure between ``s'' and ``t'' may contain arbitrary branches and connections.}
\label{FigLinear}
\end{figure}

For easy illustration, we start by solving Eqn.~\ref{eqnCostReforward} on Linear Computation Graphs (LCG) (Fig.~\ref{FigLinear} (a)). For LCGs, Eqn.~\ref{eqnCostReforward} can be solved in two cases. 

\textbf{Case(1) LCG with Identical Vertex Cost:} Suppose a LCG has $N$ vertices, each of which has identical cost as $l(v_i) = 1$ and the total cost of these $N$ vertices is $N$. Obviously, the optimal solution is reached when GCs in $V^R$ are distributed evenly in the LCG, i.e. splitting the computation graph into equal length segments. Suppose the number of vertices in $V^R$ is $k$. The total cost is then $k+\frac{N}{k}$. The optimal solution of Eqn.~\ref{eqnCostReforward} is achieved when $k=\sqrt[]{N}$, and the optimal total cost is $2\sqrt[]{N}$.

From Case(1), we can get a sense of the effectiveness of Gradient CheckPointing. The original memory cost is $N$, and can be reduced to $2\sqrt[]{N}$ at the time overhead of extra local forwards. When the network is deep, i.e. $N$ is large, huge amount of memory cost can be cut off. For example, when $N=100$, we can reduce the memory cost to $20\%$ of the original cost. Chen's algorithm \cite{chen2016training} is developed exactly from this observation and thus is only optimal in this case.

\textbf{Case (2) LCG with Non-identical Vertex Cost:}
When the assumption of identical cost does not hold, the solution to Eqn.~\ref{eqnCostReforward} does not have an analytic form. Denote the maximal Re-forward cost $\max \limits_{i} l(v_i^R, v_{i+1}^R))$ as a constant $C$, and the solution to Eqn.~\ref{eqnCostReforward} is reduced to solving for $\min \limits_{V^R} \sum_i l(v_i)$, such that all the re-forward memory costs satisfy the constraint $l(v_i^R, v_{i+1}^R)) \leq C$.


 Given a constant $C$ as the constraint, we can solve the reduced problem by constructing a new graph, called Accessibility Graph $G^A=\big(E^A, V\big)$. The edges of $G^A$, called Accessibility Edge $e_{jk}^A$, exists between vertex $v_j$ and $v_k$ if and only if $l(v_j, v_k) \leq C$, which means that $v_j$ and $v_k$ can be selected as adjacent GCs under the constraint. 
 
 Now the constraints are all encoded in the accessibility graph, we can solve the unconstrained problem $\min \limits_{V^R} \sum_i l(v_i^R)$, which is equivalent to finding the shortest path from the source vertex and the target vertex in the Accessibility Graph. Notice that in the optimal solution of Eqn.~\ref{eqnCostReforward}, $\max \limits_{i} l(v_i^R, v_{i+1}^R)) = C = l(v_j, v_k)$. $C$ would be the cost $l(v_j, v_k)$ of a vertex pair. Therefore, to determine $C$, we can simply traverse all possible $C$ by using the cost of every vertex pair, and find optimal solution under each $C$ as constraint. The best of it would then be the optimal solution of Eqn.~\ref{eqnCostReforward}. \textbf{Algorithm 1} summarizes the steps for searching an optimal solution for LCGs. For a computation graph with $|V|$ vertices and $|E|$ edges, the time complexity of \textbf{Algorithm 1} is $O(|V|^2|E| + |V|^3\log|V|)$.

\begin{algorithm}[h]  
\caption{Linear Computation Graph (LCG) Solver}  
\textbf{Input:} a linear computation graph $G$ \\
\textbf{Output:} optimal GCs $V^R$\\
\begin{algorithmic}[1]  
\FOR{each vertex pair $(v_j, v_k)$ in $G$}
\STATE Set the maximal term $C = l(v_j,v_k)$
\STATE Construct Accessibility Graph $G^A$
\STATE Find the shortest path in the Accessibility Graph as a candidate solution $V^R$
\STATE Compute the total cost of candidate solution $V^R$
\STATE Save the solution $V^R$ if the total cost is smaller.

\ENDFOR
\end{algorithmic}  
\end{algorithm}

\section{Arbitrary Computation Graph(ACG)}
As the generalization of LCGs, we present theory and algorithms for DNNs with Arbitrary Computation Graphs (ACG), in particular the acyclic directed graphs~(Fig.~\ref{FigLinear} (b)). 


\subsection{Independent Segment(IS)}
GCs break the computation graph into different segments, where we can perform re-forward and backward independently. We call it Independent Segment (IS). For LCGs, any set of GCs naturally break the computation graph into linearly arranged IS. But for ACGs, this property may not hold.

\begin{figure}[h]
\centering
\includegraphics[width=.5\linewidth]{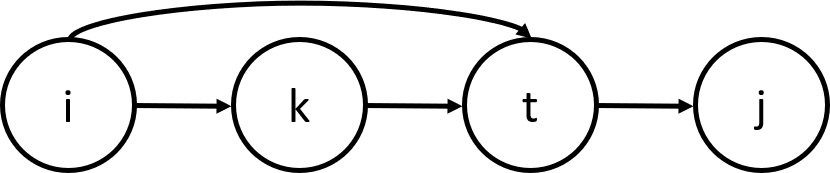}\hspace{-.1cm}
\caption{Letter represents index of the vertex in the computation graph. $v_i$ and $v_t$ can form an independent segment while $v_k$ and $v_j$ can not.}
\label{FigISFeasibility}
\end{figure}

For example, in Fig.~\ref{FigISFeasibility}, GCs $\{v_k, v_j\}$ is not feasible because $v_t$ depends on $v_i$. Thus $v_k$ and $v_j$ cannot form an IS. $v_i$ and $v_t$ can form an IS because vertices in the segment (just $v_k$ for this example) do not depend on any vertices outside the segment. Thus GCs $\{v_i, v_t\}$ is a feasible solution.

We investigate the properties of IS to better understand the solution space of GCs. Therefore, we formally define Independent Segment as following.

\begin{Def}
\textbf{Independent Segment(IS):} Independent Segment $s_{ij}=(E^{ij}, V^{ij})$ is a subgraph of the computation graph $G=(E, V)$, with $v_i$ being the source vertex and $v_j$ being the target vertex. The vertices inside $s_{ij}$ has no connections with the vertices outside $s_{ij}$, i.e. $\not\exists e_{kt}, s.t. v_k\in(V_{ij} - \{v_i, v_j\}), v_t\in(V-V_{ij})$.
\end{Def}

Given IS $s_{ij}$ and $v_i$, $v_j$ as GCs, the reforwarding and backward memory cost for this segment is the sum over cost of all the vertices inside this segment, i.e. $l(s_{ij}) = \sum_k l(v_k), v_k\in(V_{ij} - \{v_i, v_j\})$. We can then derive objective function for ACG similar as in LCG Eqn. ~\ref{eqnCostReforward}.

\begin{equation}
\min\limits_{V^R}( \sum_i l(v_i^R) + \max l(s_{ij}^R)), \label{eqnCostReforwardACG}
\end{equation}

where the second term is the maximum cost over all the IS formed by GCs in $V^R$

\subsection{IS Type and Division}

\begin{figure}[h]
\centering
\includegraphics[width=0.99\linewidth]{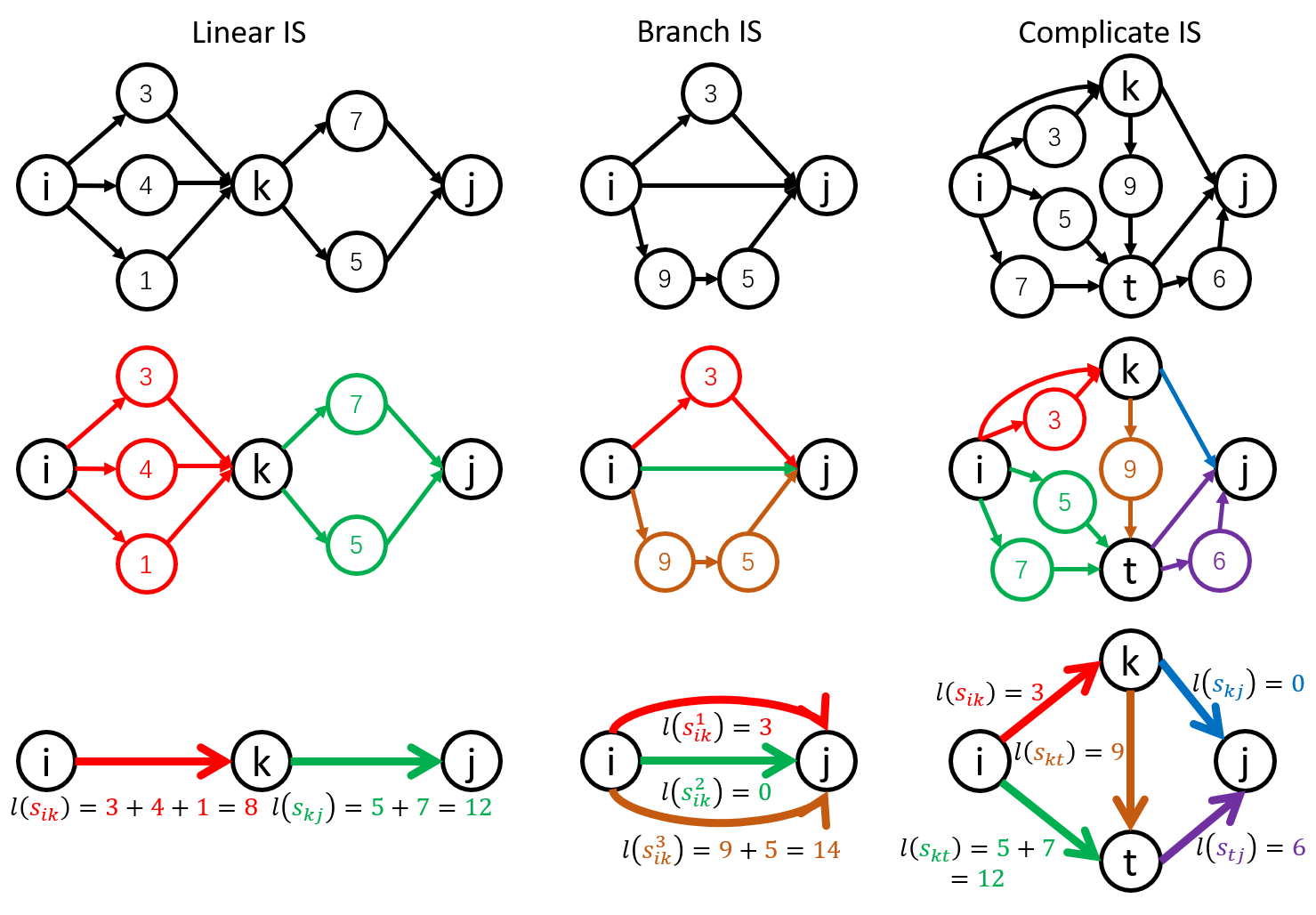}\hspace{-.1cm}
\caption{The first row is the computation graph of IS. The second row is the division of IS as colored sub-graphs. The third row is the hyper-graph of IS with the division folded into bold colored segments, associated with their memory costs. Letters represent vertex indices and numbers represent the memory cost of vertices. (Best viewed in colors.) }
\label{FigISDivision}
\end{figure}

Similar to \textbf{Algorithm 1}, we also define $C=\max l(s_{ij}^R)$ and try to solve $\min\limits_{V^R}\sum_i l(v_i^R)$ under constraint $C$. 

Given an IS $s_{ij}$, if $l(s_{ij}) > C$ and breaks constraint, we need to break down $s_{ij}$ and find more GCs inside $s_{ij}$. In other words, $s_{ij}$ is too big to satisfy constraint $C$ and we need to break it down into smaller segments so that each segment confines with the constraint.

Another question rises: how to divide an IS into a set of smaller IS? We categorize IS into different types and propose divisions for them respectively.

\begin{Def} \textbf{Linear Splitting Vertex:}
A vertex $v_t \in s_{ij}$ is a linear splitting vertex of $s_{ij}$ if and only if $s_{it}$ is valid IS, $s_{tj}$ is valid IS and $s_{ij} = s_{it} \cup s_{tj}$ and $s_{it} \cap s_{tj} = \{v_t\}$
\end{Def}

\begin{Def}  \textbf{Linear IS:}
An IS with at least one linear splitting vertex.
\end{Def}

The definition of \textbf{Linear Splitting Vertex} is to describe whether an IS can be divided into two linearly arranged IS. We categorize an IS as \textbf{Linear IS} if it has at least one linear splitting vertex. The division of \textbf{Linear IS} is naturally all of the linear segments separated by its linear splitting vertices. An example of Linear IS and its division can be viewed in the first column of Fig. ~\ref{FigISDivision}. In this example, the division of Linear IS $s_{ij}$ is $\{s_{ik}, s_{kj}\}$. If we represent member IS of the division $s_{ik}$, $s_{kj}$ as edges with memory cost, then the computation graph will be a simple linear graph.

\textbf{Algorithm 2} gets the division of Linear IS $s_{ij}$. For each vertex $v_t\in V^{ij}$, we judge whether it's a Linear Splitting Vertex with Definition 2. If it is, we break down $s_{ij}$ into two segments and try to find more Linear Splitting Vertex in each segment recursively. At the end of \textbf{Algorithm 2}, we can find all the Linear Splitting Vertices in $s_{ij}$, and get the division of $s_{ij}$ formed by IS separated by the Linear Splitting Vertices. Suppose there are $|V^{ij}|$ vertices in $s_{ij}$, the time complexity of \textbf{Algorithm 2} is $O(|V^{ij}|^3)$.
\begin{algorithm}[h]  
\caption{Get the division of Linear IS \\$\{s\}\leftarrow func(s_{ij})$}  
\textbf{Input:} IS $s_{ij}=(V^{ij}, E^{ij})$\\
\textbf{Output:} division of $s_{ij}$ (a set of IS)\\
\begin{algorithmic}[1]  

\FOR {each vertex $v_t\in V^{ij}$}
\STATE Let $\{v_{in}\}$ be the vertices of all the vertices within $s_{ij}$ that have paths to $v_t$. Let $\{v_{out}\}$ be the vertices of all the vertices within $s_{ij}$ that have paths from $v_t$.
\IF {$\{v_{in}\} \cup \{v_{out}\} \cup \{v_t\} = V^{ij}$ and $\{v_{in}\} \cap \{V_{out}\} = \emptyset$ and $\not\exists v_1\in \{v_{in}\}, v_2\in \{v_{out}\}$, $v_1, v_2$ have connections}
\STATE Return $func(s_{it})\cup func(s_{tj})$
\ENDIF
\ENDFOR

\STATE Return $\{s_{ij}\}$

\end{algorithmic}  
\end{algorithm}

\begin{Def}  \textbf{Branch IS:} an IS $s_{ij}$ with 0 linear splitting vertex and can be divided into branches: multiple IS with source vertex $v_i$ and target vertex $v_j$, i.e. $s_{ij}=s_{ij}^1 \cup s_{ij}^2 \cup ... \cup s_{ij}^n$ and $s_{ij}^1 \cap s_{ij}^2 \cap ... \cap s_{ij}^n = \{v_i, v_j\}$.
\end{Def}

For IS with no linear splitting vertex, we categorize the one formed by branches as \textbf{Branch IS}. The division of a Branch IS is simply its branches. An example of Branch IS and its division can be viewed in the second column of Fig. ~\ref{FigISDivision}. In this example, the division of Branch IS $s_{ij}$ are its branches $\{s_{ij}^1, s_{ij}^2, s_{ij}^3\}$.

\textbf{Algorithm 3} gets the division of Branch IS $s_{ij}$. If $s_{ij}$ has edge $e_{ij}$, we treat the edge $e_{ij}$ itself as a branch, add it into the division and look for more branches in the remaining graph recursively. Otherwise, we initialize an IS $s_b$ with a random vertex $v_k$, and do BFS to gradually add vertices and edges into $s_b$. If $s_{ij}$ has no branch, the edges $E^b$ of $s_b$ will end up being $E^{ij}$. Otherwise, $s_b$ will be a branch of $s_{ij}$. We add $s_b$ into division and look for more branches in the remaining graph recursively. Suppose there are $|V_{ij}|$ vertices in $s_{ij}$, the time complexity of \textbf{Algorithm 3} is $O(|V_{ij}|^2)$.

\begin{algorithm}[h]  
\caption{Get the division of Branch IS \\$\{s\}\leftarrow func(s_{ij})$}  
\textbf{Input:} IS $s_{ij}=(V^{ij}, E^{ij})$\\
\textbf{Output:} division of $s_{ij}$ (a set of IS)\\
\begin{algorithmic}[1]  
\IF {$|V_{ij} - \{v_i, v_j\}| >= 1$}
\IF {$e_{ij}\in E^{ij}$}
\STATE $s_{b} = (\{v_i, v_j\}, \{e_{ij}\})$
\STATE $s_{ij}^\prime = (V^{ij}, E^{ij}-\{e_{ij}\})$
\STATE Return $\{s_{b}\} \cup func(s_{ij}^\prime)$
\ELSE
\STATE Initialize an IS $s_b=(V^b, E^b)$. $V^b=\{v_k\}$, $v_k$ is a randomly chosen vertex in $V^{ij}- \{v_i, v_j\}$, $E^b=\emptyset$.
\STATE Initialize an empty queue $q$, add $v_k$ to $q$
\WHILE{$|q| > 0$}
\STATE pop $v_q$ from $q$
\FOR{each $v_t$ that has edge $e_{tq}$ or edge $e_{qt}$ connects to $v_q$}
\IF{$v_t \not\in V^b$ and $v_t \in (V^{ij}- \{v_i, v_j\})$}
\STATE Add $v_t$ to $q$, add $v_t$ to $V^b$, and add $e_{tq}$ or $e_{qt}$ to $E^b$
\ENDIF
\ENDFOR
\ENDWHILE
\IF {$E^b=E^{ij}$}
\STATE Return $\{s_{ij}\}$
\ELSE
\STATE $s_{ij}^\prime = (V^{ij} - V^b + \{v_i, v_j\}, E^{ij}-E^b)$
\STATE Return $\{s_{b}\} \cup func(s_{ij}^\prime)$
\ENDIF
\ENDIF
\ELSE
\STATE Return $\{s_{ij}\}$
\ENDIF
\end{algorithmic}  
\end{algorithm}

\begin{Def} \textbf{Complicate IS:}
A Complicate IS is an IS having 0 linear splitting vertex and 0 branch.
\end{Def}

For the remaining IS with no linear splitting vertex and no branch, we categorize it as \textbf{Complicate IS}, because it's not very straight forward to get the division of this type of IS. For Complicate IS, we don't want a trivial division such that each member IS is formed by a single vertex. Instead, we want the member IS is as large as possible. Therefore, we define the division of Complicate IS as following.

\begin{Def} \textbf{Division of Complicate IS:}
$\{s_{pq}\}$ is the division of Complicate IS $s_{ij}$. For each member IS $s_{pq}\in\{s_{pq}\}$, there doesn't exist another $s_{kt}$, such that $s_{pq}\subsetneqq s_{kt} \subsetneqq s_{ij}$
\end{Def}

We prove that the Division of Complicate IS is unique and details of proof can be viewed in supplementary material. An example of Complicate IS and its division can be viewed in the third column of Fig. ~\ref{FigISDivision}. In this example, the division of Complicate IS $s_{ij}$ are $\{s_{ik}, s_{it}, s_{kt}, s_{kj}, s_{tj}\}$. For any member IS, $s_{ik}$ for example, there cannot exist an IS in $s_{ij}$ that can contain it.

\textbf{Algorithm 4} gets the division of Complicate IS $s_{ij}$. First we get all the possible IS within $s_{ij}$ and put them into a set $S$. Then for each IS $s_{kt}\in S$, if there is no other IS $s_{ab}\in S$ containing $s_{kt}$, i.e. $s_{kt}$ is already as large as possible, we put $s_{kt}$ into the division. Suppose there are $|V^{ij}|$ vertices and $|E^{ij}|$ edges in $s_{ij}$, the time complexity of \textbf{Algorithm 4} is $O(|V^{ij}|^2|E^{ij}|+|V^{ij}|^3)$.

\begin{algorithm}[h]  
\caption{Get the Division of Complicate IS $s_{ij}$\\$\{s\}\leftarrow func(s_{ij})$}  
\textbf{Input:} IS $s_{ij}=(V^{ij}, E^{ij})$\\
\textbf{Output:} the Division of $s_{ij}$(a set of IS)\\
\begin{algorithmic}[1]  
\STATE Initialize an empty IS set $S=\emptyset$.
\STATE Initialize an empty set $D=\emptyset$ for division.
\FOR {each vertex pair $(v_k, v_t)$ except $(v_i, v_j)$ in $s_{ij}$}
\STATE For all the vertices $\{v\}$ that have paths from $v_k$ and have paths to $v_t$.
\IF {$\not\exists v_p \not\in \{v\} \cup \{v_k, v_t\}$, $v_p$ has connection to a $v_q \in \{v\}$}
\STATE Vertex $v_k$ and $v_t$ can form an IS. Add IS $s_{kt}$ to $S$
\ENDIF
\ENDFOR
\FOR {each IS $s_{kt}\in S$}
\STATE If there doesn't exist a $s_{ab}\in S$ such that $s_{kt} \subsetneqq s_{ab} \subsetneqq s_{ij}$, put $s_{kt}$ into $D$.
\ENDFOR
\STATE Return $D$
\end{algorithmic}  
\end{algorithm}

\subsection{Division Tree and ACG Solver}
\begin{figure}[h]
\centering
\includegraphics[width=0.85\linewidth]{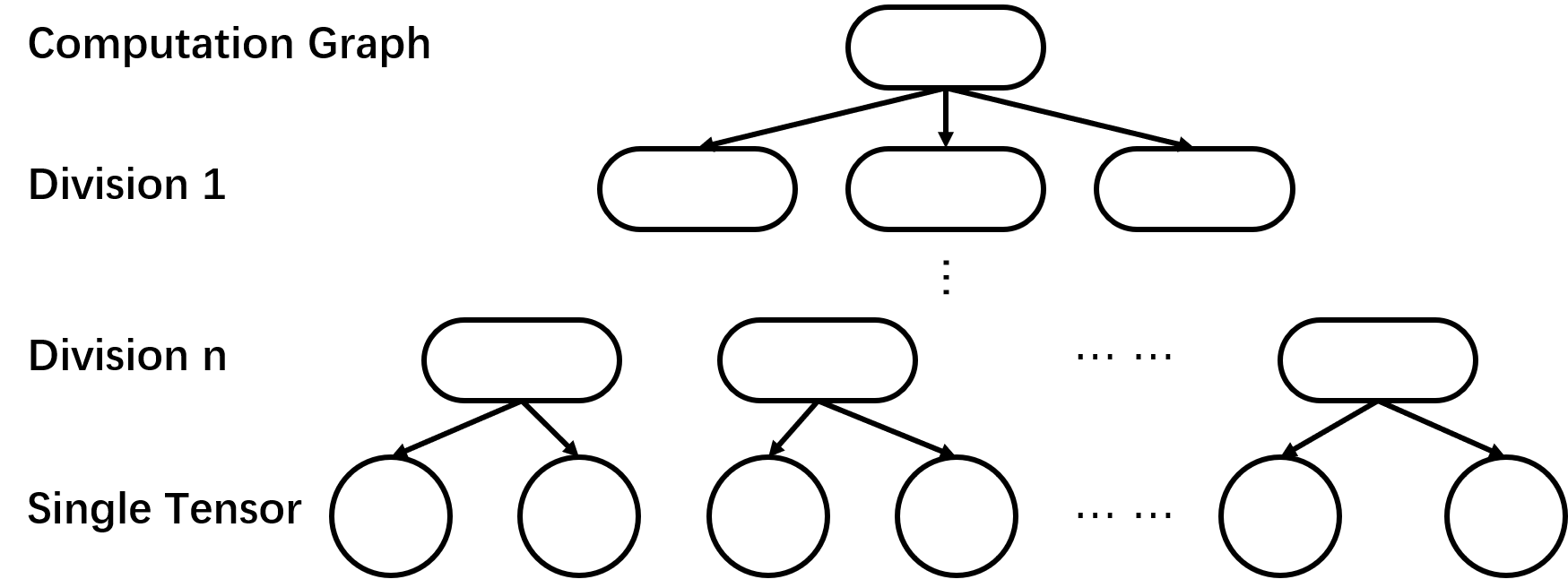}\hspace{-.1cm}
\caption{Division tree of a computation graph. The root node is the whole computation graph (largest IS). All the leaf nodes are single tensors (smallest IS). Children of a non-leaf node are the member IS in its division.}
\label{FigDivisionTree}
\end{figure}

With the definition of three types of IS and their divisions, we can build a division tree from the computation graph (Figure~\ref{FigDivisionTree}) where a non-leaf node would be an IS and its children would be its corresponding division. The root node of the whole computation graph is the largest IS, and the leaf nodes are single tensors in the computation graph. 

\begin{Theo}
The division tree of a computation graph is unique and complete.
\end{Theo}

We prove that the division tree of a computation graph is unique and complete. The uniqueness indicates that an ACG can only have one division tree. The completeness indicates that the division tree represents the whole solution space for optimal GCs searching, which means finding the optimal solution in division tree is equivalent to finding the optimal solution in computation graph.

With the division tree, we can search for optimal GCs recursively. The recursion starts at the biggest IS (the whole computation graph, root node of division tree) and ends at the smallest IS (single vertex). 

\textbf{Algorithm 5} tries to find optimal GCs in the division tree recursively given constraint $C$. The main idea is that for an IS, if its cost satisfies the constraint $C$, there's no need to find more GCs inside the IS. If an IS breaks the constraint $C$, then we will have to find more GCs inside the segment. For Branch IS and Complicate IS, once it's broken down, we add the connecting vertices of all its member IS to GCs. For Branch IS $s_{ij}$, the connecting vertices are simply $v_i$ and $v_j$. For Complicate IS, it can be more. For example, in third column of Fig. ~\ref{FigISDivision}, the connecting vertices are $v_i, v_k, v_t, v_j$.

\begin{figure}[h]
\centering
\includegraphics[width=0.99\linewidth]{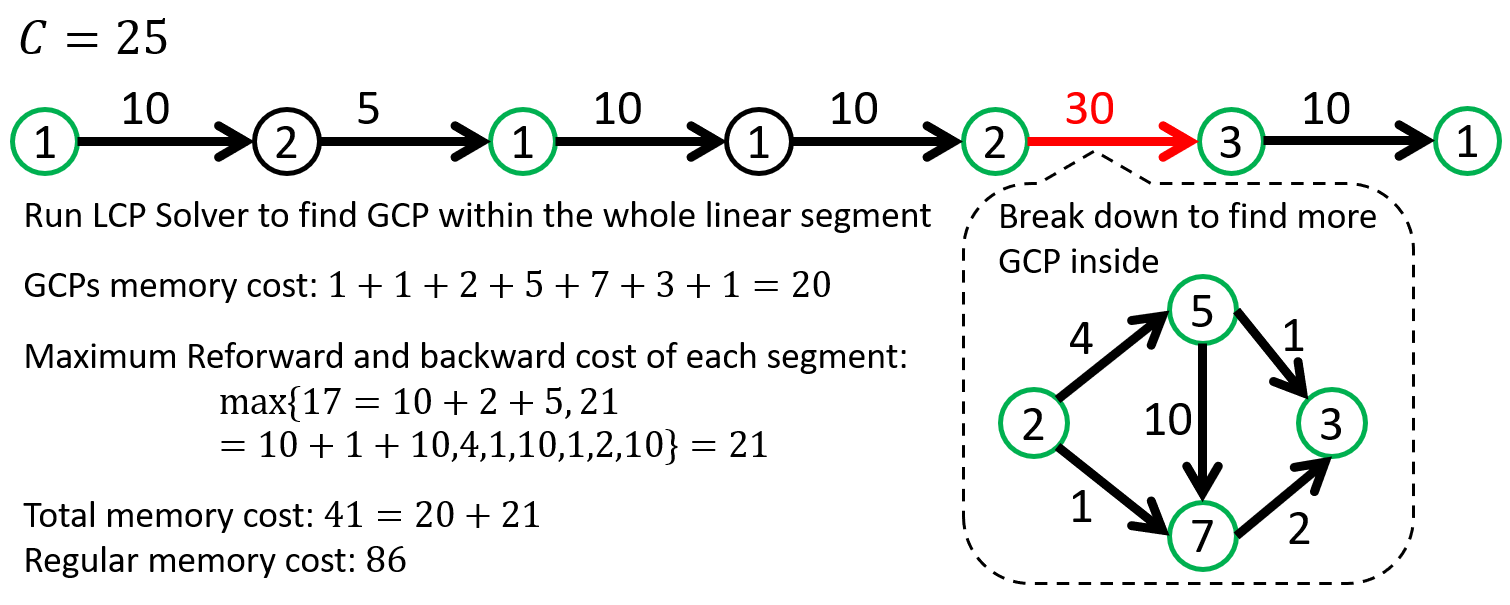}\hspace{-.1cm}
\caption{Running LCG Solver on Linear IS.}
\label{FigDivisionTreeRecur}
\end{figure}

For Linear IS, it can be redundant to add all the connecting vertices of member IS into GCs. For example, in Fig.~\ref{FigDivisionTreeRecur}, given $C=25$, the cost of member IS is $10,5,10,10,30,10$ respectively. The member IS with cost $30$ needs to be further broken down to find more GCs inside. And it splits the whole graph into two linear graphs (folding the member IS as edge). For the linear graph on the left, we can further run our LCG Solver to find optimal GCs out of the vertices.

\begin{algorithm}[h]  
\caption{Find GCs in division tree recursively given constraint $C$:\\$V^R\leftarrow recur(s, V^R, C)$}  
\textbf{Input:} an IS $s_{ij}$, current GCs $V^R$, max term $C$ \\
\textbf{Output:} new GCs $V^R$ with GCs inside $s_{ij}$ added\\
\begin{algorithmic}[1]  
\IF {cost of $s$ smaller or equal than $C$}
\STATE Return $V^R$
\ENDIF

\IF {$s_{ij}$ is Linear IS}
\STATE Sort the division topologically. Suppose division of $s_{ij}$ is $\{s_{12}, s_{23}, ..., s_{(n-1)n}\}$, and $v_1=v_i$, $v_n=v_j$. 
\STATE Set starting index $p=1$
\FOR{each member IS $s_{kt}$ in the division}
\IF {cost of $s_{kt}$ breaks constraint: $l(s_{kt}) > C$}
\STATE Build a linear graph $G^\prime$, with vertices as $\{v_p, v_{p+1}, ..., v_{k}\}$, and edges as $\{e_{p(p+1)}, ... e_{(k-1)k}\}$, with costs of edges as $\{l(s_{p(p+1)}), ..., l(s_{k-1)k})\}$
\STATE Solve $G^\prime$ with LCG Solver with constraint $C$ and add GCs to $V^R$: $V^R=V^R+LCGSolver(G^\prime, C)$
\STATE$ V^R = recur(s_{kt}, V^R, C)$
\ENDIF
\ENDFOR
\ELSE
\FOR{each member IS $s_{kt}$ in the division of $s_{ij}$}
\STATE Add $v_k$ and $v_t$ to $V^R$
\IF {cost of $s_{kt}$ breaks constraint: $l(s_{kt}) > C$}
\STATE $V^R=recur(s_{ij}, V^R, C)$
\ENDIF
\ENDFOR
\ENDIF
\STATE Return $V^R$

\end{algorithmic}  
\end{algorithm}

Finally we can put together our ACG Solver (\textbf{Algorithm 6}) with all the components we have discussed before. First we get a list of possible max term $C$ from all possible IS in the computation graph, similar to the LCG Solver. Then we build the division tree with \textbf{Algorithm 2-4}. For each max term $C$, the recursion function in \textbf{Algorithm 5} is called with the whole computation graph (the largest IS) as the input. The optimal GCs set is the one with the lowest overall memory cost across all the max terms $C$. Suppose there are $|V|$ vertices and $|E|$ edges in computation graph, the overall time complexity of \textbf{Algorithm 6} is $O(|V|^2|E| + |V|^3\log|V|)$. Note that given an ACG, ACG Solver is a pre-procession step and only needs to run once before the Gradient CheckPointing training.

\begin{algorithm}[h]  
\caption{Arbitrary Computation Graph (ACG) Solver}  
\textbf{Input:} an arbitrary computation graph $G$ \\
\textbf{Output:} optimal GCs $V^R$\\
\begin{algorithmic}[1]  
\STATE Get all possible IS and their costs. Use their costs to form the max term list $\{c\}$.
\STATE Build the division tree from computation graph: from the root node (the computation graph), build its children from its division, until all leaf nodes are single tensors.
\FOR{each possible max term $C$ in max term list $\{c\}$}
\STATE Set $V^R$ empty
\STATE $V^R=recur(G, V^R, C)$
\STATE Summarize the total loss, save the current solution $V^R$ if it's better.
\ENDFOR
\STATE Return $V^R_{best}$
\end{algorithmic}  
\end{algorithm}

\section{Experiment}

\begin{table*}
\caption{Training memory cut-offs and time overheads of GCP training with respect to regular training. The GCs used in GCP training are provided by Random (baseline), Chen's~\cite{chen2016training} and our GC algorithm, respectively. Note that the "Random" baseline reports the \textbf{best} number over 10 random trials. \cite{chen2016training} is \textbf{not applicable} to non-linear networks (or ACGs). The random strategy is \textbf{not applicable} to three networks from NAS because it cannot find a valid solution after 100 trials. Although using different GCs from random, Chen's and our GC algorithm, GCP training always conducts one extra forwarding, therefore share the same ``GCP Time'' and  ``GCP Time Overhead'' columns in the table.}
\label{TabResult}
\centering
\fontsize{8}{10}\selectfont
\begin{tabular}{|c||c|c|c|c|c||c|c|c|c|}
\hline
Linear network&\begin{tabular}{c} Regular\\Memory\\(MB)\end{tabular} &  \begin{tabular}{c} Random\\Memory\\(MB)$\downarrow$ \end{tabular} &\begin{tabular}{c} Chen's\cite{chen2016training} \\Memory\\(MB)$\downarrow$ \end{tabular}&  \begin{tabular}{c} Ours\\Memory\\(MB)$\downarrow$ \end{tabular} &\begin{tabular}{c} Ours\\Memory\\Cut-offs$\uparrow$ \end{tabular} &\begin{tabular}{c} Regular\\Time\\(Sec)\end{tabular} &\begin{tabular}{c} GCP\\Time\\(Sec)$\downarrow$ \end{tabular} &\begin{tabular}{c} GCP\\Time \\Overhead$\downarrow$ \end{tabular} \\
\hline
Alexnet batch 1024&4955&4408&4408&\textbf{3287}&34\%&0.388&0.519&34\%\\
Vgg11 batch 64&3577&\textbf{2781}&\textbf{2781}&\textbf{2781}&22\%&0.266&0.356&34\%\\
Vgg13 batch 64&5136&\textbf{3565}&\textbf{3565}&\textbf{3565}&31\%&0.418&0.558&33\%\\
Vgg16 batch 64&5136&4352&3957&\textbf{3565}&31\%&0.503&0.666&32\%\\
Vgg19 batch 64&5189&4548&3957&\textbf{3565}&31\%&0.581&0.774&33\%\\
\hline
Non-linear network&\begin{tabular}{c} Regular \\Memory\\ (MB) \end{tabular} &\begin{tabular}{c} Random\\Memory\\(MB)$\downarrow$ \end{tabular} &\begin{tabular}{c} Chen's\cite{chen2016training} \\Memory\\ (MB)$\downarrow$ \end{tabular}&    \begin{tabular}{c} Ours\\Memory\\(MB) $\downarrow$\end{tabular}&\begin{tabular}{c} Ours\\Memory\\Cut-offs$\uparrow$ \end{tabular}  &\begin{tabular}{c} Regular\\Time\\(Sec) \end{tabular} &\begin{tabular}{c} GCP\\Time\\(Sec)$\downarrow$\end{tabular}&\begin{tabular}{c} GCP\\Time \\Overhead$\downarrow$\end{tabular}\\
\hline
ResNet18 batch 256&5635&4069&N/A&\textbf{3677}&35\%&0.422&0.548&30\%\\
ResNet34 batch 128&4079&2231&N/A&\textbf{1838}&55\%&0.364&0.493&35\%\\
ResNet50 batch 64&5323&2714&N/A&\textbf{1973}&63\%&0.394&0.516&31\%\\
ResNet101 batch 32&3934&2541&N/A&\textbf{1024}&74\%&0.356&0.482&35\%\\
ResNet152 batch 16&2767&1464&N/A&\textbf{526}&81\%&0.241&0.331&37\%\\
Densenet121 batch 32&4027&1629&N/A&\textbf{898}&78\%&0.218&0.292&34\%\\
Densenet161 batch 16&3751&1432&N/A&\textbf{666}&82\%&0.252&0.341&36\%\\
Densenet169 batch 32&4862&1774&N/A&\textbf{897}&82\%&0.270&0.357&32\%\\
Densenet201 batch 16&3146&1242&N/A&\textbf{474}&85\%&0.200&0.306&53\%\\
Inceptionv3 batch 32&3074&1336&N/A&\textbf{881}&71\%&0.291&0.374&29\%\\
NASNet batch 64&5832&N/A&N/A&\textbf{1129}&81\%&0.408&0.535&31\%\\
AmoebaNet batch 64&4944&N/A&N/A&\textbf{1058}&79\%&0.331&0.450&36\%\\
DARTS batch 64&5627&N/A&N/A&\textbf{1115}&80\%&0.318&0.494&55\%\\
\hline
\end{tabular}
\end{table*}

We evaluated our approach on (1) networks with LCGs, such as Alexnet \cite{krizhevsky2012imagenet} and Vgg \cite{simonyan2014very}. (2) networks with non-linear computation graphs, such as ResNet \cite{he2016deep}, Densenet \cite{huang2017densely} and Inception net \cite{szegedy2016rethinking}, and three highly non-linear networks from NAS, NASNet\cite{zoph2018learning}, AmoebaNet~\cite{real2019regularized} and DARTS~\cite{liu2018darts}. In Table~\ref{TabResult}, We compared our approach with Chen's algorithm \cite{chen2016training} and a random baseline and the regular training approach. \textbf{Note that Chen's algorithm only works on LCGs and is not applicable to non-linear computation graphs}. Our approach directly works on arbitrary computation graphs. For random baseline, we randomly select 1-5 GCs among all vertices in the computation graph. We repeat this random selection for 10 times and report the \textbf{best} solution (i.e. the solution with minimal memory consumption) among 10 trials. For non-linear networks, random selection can yield invalid solution (unable to do independent forward and backward between GCs). In this case, we repeat random selection process until we have 10 valid solutions and report the \textbf{best} results among them.

All experiments were conducted in Pytorch~1.5. GPU memory costs (MB) are measured in Float32. The reported memory costs have excluded the stationary cost, such as model weights and Pytorch CUDA interface. The input to Inceptionv3 is $[BatchSize, 3, 300, 300]$, the input to three NAS networks is $[BatchSize, 3, 32, 32]$, and the input to all other networks is $[BatchSize, 3, 224, 224]$. Although using different GCs from the random, Chen's and our GC algorithm, GCP training always conducts one extra forwarding, therefore costs the same ``GCP Time'' and  ``GCP Time Overhead'' in Table.~\ref{TabResult}. We report the GCP training time per iteration (in seconds "Sec") averaged over $100$ iterations.

Table.~\ref{TabResult} shows that our approach cuts down the most amount of memory from the regular approach. For instance, for linear network Vgg19, $31\%$ memory was cut down, enabling the GCP training that costs $33\%$ time overhead. Due to our optimal GC solution on computation graphs, GCP training using our GCs outperforms Chen's approach and also constantly outperforms the best solution of 10 random trials. For non-linear networks, Chen's approach does not apply, while our approach can still give substantial memory cut and constantly outperform the best solution of 10 random trials. On the deepest ResNet (ResNet152), $81\%$ memory cut was achieved, enabling the GCP training that costs only $37\%$ time overhead. For Densenet series and networks from NAS, more than $80\%$ memory cut were achieved with around $40\%$ time overhead. 

\section{Conclusion}
Gradient CheckPointing (GCP) is a fundamental training approach that makes it possible to train very heavy DNNs on finite GPU memory. Automatic Gradient Checkpoint(GC) searching is the key to GCP, whereas existing efforts are stagnant at heuristic GC searching and LCGs. To our knowledge, our theoretical and algorithmic results are the first top-down work that achieves an optimal memory GC solution for DNNs with arbitrary computation graphs. 
Our advance of GCP is general and can be further integrated with any low-level techniques such as distributed computing, GPU/CPU swapping, computation graph optimization and liveness analysis.

{\small
\bibliographystyle{ieee_fullname}
\bibliography{egbib}
}

\end{document}